\documentclass{article}

\usepackage[preprint,nonanonymous]{neurips_2026}

\usepackage[utf8]{inputenc} % allow utf-8 input
\usepackage[T1]{fontenc}    % use 8-bit T1 fonts
\usepackage{hyperref}       % hyperlinks
\usepackage{url}            % simple URL typesetting
\usepackage{booktabs}       % professional-quality tables
\usepackage{amsfonts}       % blackboard math symbols
\usepackage{nicefrac}       % compact symbols for 1/2, etc.
\usepackage{microtype}      % microtypography
\usepackage{xcolor}         % colors
\usepackage{graphicx}
\usepackage{wrapfig}
\usepackage{capt-of}

\newcommand{\vseq}{\mathbf{v}}
\newcommand{\aseq}{\mathbf{a}}

\usepackage{amsmath}
\usepackage{algorithm}
\usepackage{algpseudocode}
\newcommand{\methodname}{NoiseGate}

\title{NoiseGate: Learning Per-Latent Timestep Schedules\\
  as Information Gating in World Action Models}

\author{%
  \textbf{Wen Huang$^{1,\dagger}$\thanks{$^{\dagger}$Equal contribution (co-first authors).}, Haoran Sun$^{2,\dagger}$, Yongjian Guo$^{1,\dagger}$, Yunxuan Ma$^{2}$, Haoran Li$^{2,3}$, Jing Long$^{2,3}$} \\
  \textbf{Zhouying Mo$^{4}$, Zhong Guan$^{4}$, Yucheng Guo$^{3}$, Shuai Di$^{3}$, Junwu Xiong$^{3}$} \\
  $^{1}$Tsinghua University, $^{2}$Peking University, $^{3}$JDT AI Infra, $^{4}$Tianjin University \\
  \texttt{\{huang-w24, guo-yj24\}@mails.tsinghua.edu.cn, sunhaoran0301@stu.pku.edu.cn}
}

\begin{document}

\maketitle

\begin{abstract}

World Action Models (WAMs) are an emerging family of policies that tie robot action generation to future-observation modeling. In this work, we focus on the joint video--action modeling paradigm, where actions and imagined future observations are co-generated along a shared denoising or flow trajectory, so that perception, prediction, and control are coupled within one generative process.
Existing WAMs typically realize this paradigm with a Mixture-of-Transformers (MoT), where video and action tokens interact through shared self-attention. This architecture can in principle assign a separate timestep $t_f$ to each predicted latent frame, yet current systems collapse this degree of freedom onto a single shared scalar $t$. Under the noise-as-masking view of Diffusion Forcing, this shared schedule imposes the unjustified prior that every predicted latent is equally reliable for action generation. We instead view the per-latent schedule as a \emph{learnable information-gating policy}: by changing a latent frame's noise level, the policy modulates the reliability of its Key/Value contribution to the action tokens. We propose \textbf{NoiseGate}, which combines independent per-latent timestep sampling during backbone training, a lightweight Gating Policy Network that emits per-latent time increments during denoising, and task-reward optimization that trains the schedule policy without hand-crafted shape priors. Built on a joint video--action MoT backbone, NoiseGate delivers consistent gains on diverse RoboTwin random-scene manipulation tasks.

\end{abstract}
\section{Introduction}
\label{sec:intro}

\begin{wrapfigure}{r}{0.5\linewidth}
  \vspace{-0.8em}
  \includegraphics[width=0.98\linewidth]{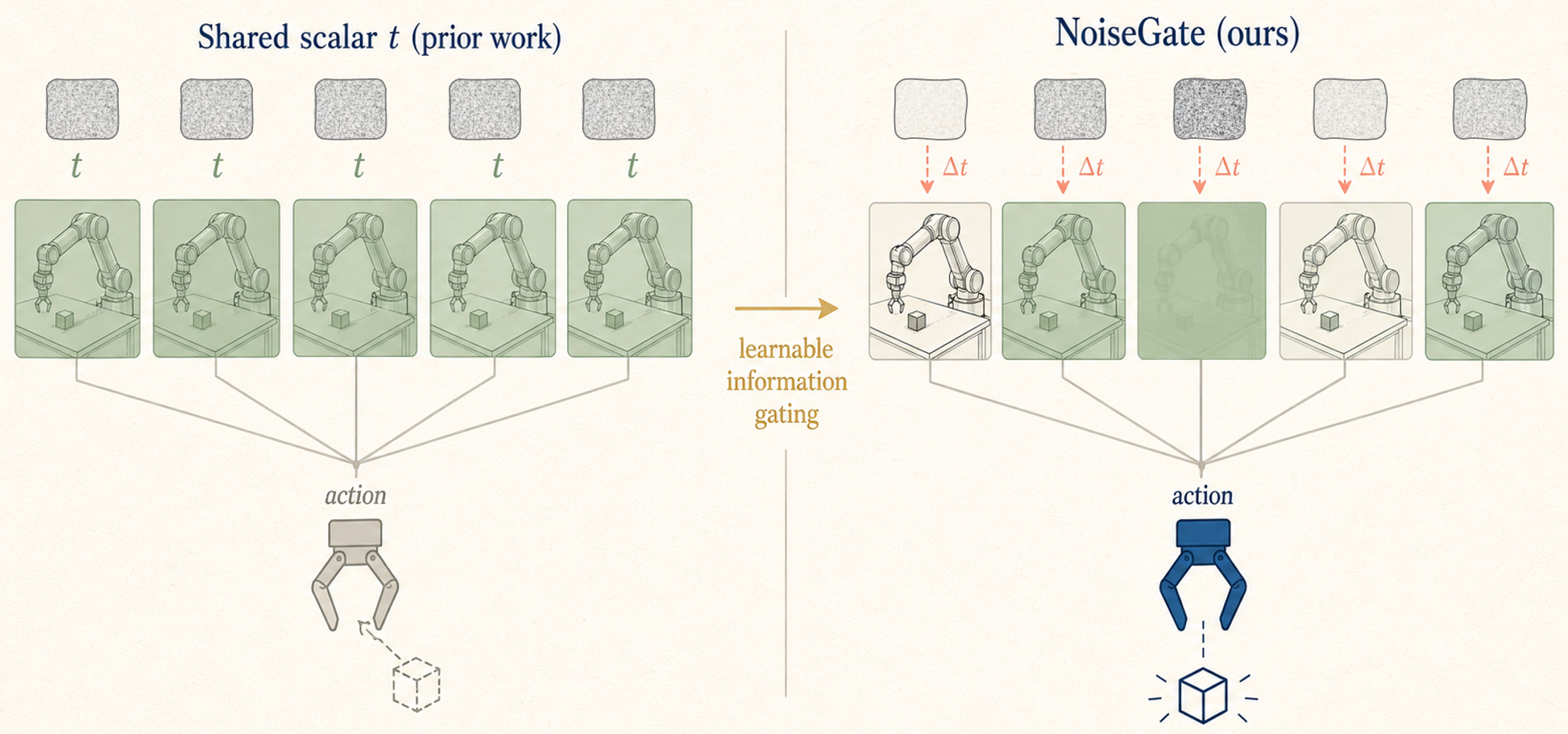}
  \caption{\textbf{Method preview.} \methodname{} learns per-latent schedules as task-adaptive information gates in a joint video-action denoising backbone.}
  \label{fig:preview}
  \vspace{-0.8em}
\end{wrapfigure}

World Action Models (WAMs)~\cite{bi2025motus, yuan2026fast, ye2026world, shen2025videovla, kim2026cosmos} generalize classical video-language-action policies by modeling the joint distribution $p(\mathbf{v}, \mathbf{a} \mid v_0,l)$ over a chunk of $F$ future latent frames $\mathbf{v} = (v_1, \dots, v_F)$ and an action chunk $\mathbf{a}$, given the current observation latent $v_0$ and language instruction $l$.
Among the various WAM designs~\cite{bi2025motus, yuan2026fast, cen2025rynnvla, li2025unified}, one prominent and effective family jointly denoises future latents and actions through a diffusion or flow-matching backbone.
This paradigm, exemplified by recent systems such as Motus~\cite{bi2025motus} and Fast-WAM~\cite{yuan2026fast}, unifies perception, prediction, and control along a common denoising trajectory, offering conceptual simplicity and strong empirical efficiency.

The backbones used by common joint video--action WAMs (e.g., chunk-bidirectional Wan~2.2~\cite{wan2025}) are in principle agnostic to how noise levels are distributed across predicted latents: each future latent $v_f$ could in principle carry its own timestep $t_f$.
Yet standard practice~\cite{bi2025motus, yuan2026fast} forces every predicted latent frame to share a single scalar $t$, collapsing the per-latent timestep vector $\mathbf{t} \in \mathbb{R}^F$ to a scalar at every denoising step.
Diffusion Forcing~\cite{chen2024diffusion} offers a useful lens: since noise level is a form of partial masking, a shared scalar $t$ encodes the strong prior that every predicted latent frame is equally ``visible'' to the action generation process.
This prior is unmotivated, because the influence of each $v_f$ on predicting the correct action is heterogeneous, task-dependent, and unknown in advance.
For example, when the model is uncertain about upcoming critical events, such as grasping or placing, it may be beneficial to keep the corresponding imagined latents blurrier.
Any handcrafted shape (e.g., monotonic in the latent-frame $f$) merely swaps one prior for another, leading to suboptimal action generation.

In this work, we move beyond the fixed shared schedule and reframe the per-latent timestep assignment as a \emph{learnable information-gating policy}.
A natural realization of the joint video--action modeling paradigm couples a video DiT with an action-expert DiT through a Mixture-of-Transformers (MoT)~\cite{liang2024mixture}: video and action tokens share a single self-attention layer, while each modality retains its own feed-forward expert.
This joint self-attention makes the gating interpretation concrete: the noise level on a latent frame directly modulates the reliability of its Key/Value contribution at every denoising step, so $\mathbf{t}$ acts as a bank of continuous gates controlling how much evidence from each future latent propagates into the action tokens (see Fig.~\ref{fig:preview}).
The optimal gating pattern is inherently task-dependent and should be learned from data rather than prescribed by a fixed schedule.

We instantiate this view as \methodname{} with three coupled ingredients.
First, during training, we sample timesteps independently per latent, borrowing the recipe of Diffusion Forcing~\cite{chen2024diffusion} but applying it to a chunk-bidirectional (non-causal) backbone, so that the model can handle arbitrary per-latent timestep profiles at inference.
Second, a lightweight \emph{Gating Policy Network} (GPN) reads the current latents and their timesteps at each denoising step and emits per-latent increments $\Delta t_f$ for the future latents $v_{1:F}$.
Third, we train the GPN with GRPO~\cite{shao2024deepseekmath} against a sparse task-success reward, grounding the gating policy in actual task utility rather than any hand-crafted prior.

Our contributions are summarized as follows:

\begin{itemize}
\item We reframe per-latent timestep scheduling in joint video--action WAMs as a learnable information-gating policy over the shared self-attention, and show that the conventional shared-scalar schedule is a strong implicit prior (\S\ref{sec:problem}).
\item We propose \methodname{}, which realizes this view via independent per-latent timestep sampling, a per-step GPN, and GRPO optimization against task reward, requiring no hand-crafted shape prior (\S\ref{sec:method}).
\item On RoboTwin under random-scene conditions, \methodname{} outperforms both shared-scalar and hand-crafted per-latent schedules, confirming that learning the gating policy is essential to exploit the per-latent degree of freedom (\S\ref{sec:experiments}).
\end{itemize}

\section{Related Work}
\label{sec:related}

\textbf{Video--action generative policies.}
Diffusion-based policy learning, beginning with Diffusion Policy~\cite{chi2025diffusion}, has been scaled to large visual backbones and language-conditioned VLA systems~\cite{shukor2025smolvla, li2025cogvla, bjorck2025gr00t, intelligence2025pi_, liu2024rdt, kim2024openvla, black2024pi_0, zitkovich2023rt, team2025gemini, shi2025memoryvla}. A recent line of World Action Models (WAMs)~\cite{bi2025motus, yuan2026fast, ye2026world, cheang2024gr, jang2025dreamgen, agarwal2025cosmos, wu2023unleashing, zhou2024robodreamer, shen2025videovla, hu2024video, li2025unified} goes beyond action-only prediction by co-generating future observations and actions, modeling $p(\mathbf{v},\mathbf{a}\mid v_0,l)$ rather than only $p(\mathbf{a}\mid v_0,l)$. We focus on the joint video--action denoising/flow-matching variant, exemplified by Motus~\cite{bi2025motus} and Fast-WAM~\cite{yuan2026fast}, where a video DiT and an action expert interact through shared self-attention in a Mixture-of-Transformers architecture~\cite{liang2024mixture}. This design exposes a rarely studied degree of freedom: each predicted latent frame could carry its own diffusion time, yet existing systems usually collapse the whole predicted chunk to one shared scalar timestep.

\textbf{Per-token noise levels and adaptive diffusion schedules.}
Diffusion Forcing~\cite{chen2024diffusion} provides the closest training-time precedent for departing from a shared timestep: it interprets noise level as partial masking and trains causal sequence models with independently sampled per-token noise levels. We borrow this noise-as-masking view and the independent noise sampling recipe, but use them for a different purpose. Our backbone is chunk-bidirectional rather than causal, and independent noise is not used as a rollout mechanism; it is a substrate that makes arbitrary per-latent noise profiles valid at inference time. Classical samplers such as DDIM~\cite{song2020denoising} and EDM~\cite{karras2022elucidating} improve the global denoising trajectory, while learned schedule methods such as TPDM~\cite{ye2025schedule} predict scalar timesteps for single-modality image generation. In contrast, our scheduler emits a vector $\mathbf{t}=(t_1,\ldots,t_F)$ over future video latents inside a joint video--action backbone, where each $t_f$ controls the reliability of that latent's Key/Value contribution to action tokens.

\textbf{Reinforcement learning for VLA policies.}
Recent work~\cite{zang2025rlinf, yu2025rlinf, guan2026rl, li2025simplevla, liu2025can, liu2025flow, zhang2025reinflow, chen2025pirl} has also shown that reinforcement learning can substantially improve VLA policies after supervised imitation. Frameworks such as RLinf-vla~\cite{zang2025rlinf} and SimpleVLA-RL~\cite{li2025simplevla} scale outcome-driven RL, including PPO- or GRPO-style optimization, to large VLA models across simulated manipulation benchmarks and real-robot settings. These methods optimize the action-producing policy itself, typically improving the model's ability to map observations and language instructions to successful action trajectories. Our use of RL is complementary: the pretrained video--action backbone is frozen, and reward optimization is applied only to a lightweight scheduler that controls how future latent frames are revealed to the action tokens during denoising. Thus the learned object is not the VLA action policy directly, but a per-latent information-gating policy inside the WAM's generative process.

Taken together, prior WAMs establish joint video--action generation, Diffusion Forcing motivates independent noise levels as masking, learned diffusion schedulers show that timestep selection can be optimized, and VLA-RL methods demonstrate the value of task-reward fine-tuning for embodied policies. \methodname{} combines these threads in a different setting: it learns a task-reward-optimized, per-latent timestep schedule for a joint video--action denoising backbone.

% ── 3. Problem Formulation ────────────────────────────────────────────────────
\section{Problem Formulation}
\label{sec:problem}

This section establishes the notation for video--action generation (\S\ref{subsec:prelim}), identifies how the conventional shared-scalar schedule limits the model's ability to handle intra-chunk information density (\S\ref{subsec:gating}), and formulates the per-frame latent scheduling problem (\S\ref{subsec:policy}).

\subsection{Preliminaries: Video--Action Generation in WAMs}
\label{subsec:prelim}

\textbf{Notation and setup.}
Let $\mathbf{v} = \{v_1, \dots, v_F\} \in \mathcal{V}^{F}$ denote a chunk of $F$ future latent frames. We take $v_0$ as the clean current observation, $l$ as the language instruction, and $\mathbf{a} \in \mathcal{A}^H$ as the action chunk co-generated with the video. The model captures the conditional distribution $p_\theta(\mathbf{v}, \mathbf{a} \mid v_0,l)$.

\textbf{Joint video--action denoising backbone.}
Current WAMs typically use a diffusion or flow-matching Mixture-of-Transformers~\cite{liang2024mixture} to jointly denoise future latents and action tokens. While the architecture nominally admits an independent timestep for every token, standard practice collapses the per-element times into two global scalars:
\begin{equation}
\label{eq:shared-t}
    g_\theta\!\left(\mathbf{v}^{\mathbf{t}_v},\,\mathbf{a}^{t_a},\, \mathbf{t}_v, t_a, v_0,l\right),
    \quad \text{where} \quad
    \mathbf{t}_v = [t_1, \ldots, t_F] \in [0,T]^F, \quad t_a \in [0,T].
\end{equation}
Crucially, in existing WAMs, even the video timesteps are collapsed such that $t_1 = \dots = t_F = t$. We re-examine this choice, specifically focusing on the flexibility of $\mathbf{t}_v$ while maintaining a synchronous schedule for action $\mathbf{a}$ to ensure a stable execution trajectory.

\subsection{Per-Latent Timesteps as Information Gates}
\label{subsec:gating}

\textbf{The Hidden Prior of a Shared Scalar $t$.}
Setting all $t_f = t$ imposes a strong prior: every future frame and every action token is assumed to be equally ``visible'' or ``reliable'' at any given step $k$. However, due to temporal causal dependencies and varying task complexity, certain future frames are inherently harder to predict or more critical for action grounding than others. 

\textbf{Video as a Gated Memory for Action.}
In the MoT backbone, video and action tokens interact through shared self-attention. Following the ``noise-as-masking'' interpretation~\cite{chen2024diffusion}, a higher noise level $t_f$ renders the $f$-th latent frame less reliable, effectively "gating" its contribution to the action tokens' hidden representations. By allowing $\mathbf{t}_v$ to be heterogeneous, we allow the model to \emph{prioritize} the denoising of specific frames that are most informative for the current action, without needing to alter the action's own denoising pace $t_a$.

\subsection{Learning the Per-Frame Schedule as a Policy}
\label{subsec:policy}

We cast the video scheduling as a policy $\pi_\phi$. At each step $k$, the policy observes the current state and determines the individual time increments for latent frames only:
\begin{equation}
    \Delta\mathbf{t}_v^{(k)} = \pi_\phi\!\left(\mathbf{v}^{\mathbf{t}_v^{(k)}}, \mathbf{t}_v^{(k)};\; v_0\right).
\end{equation}
The action schedule $t_a$ follows a fixed linear or cosine decay, acting as a global ``clock,'' while $\mathbf{t}_v$ adapts its trajectory to maximize the task-success reward $R$.

% ── 4. Method ────────────────────────────────────────────────────────────────
\section{\methodname: Per-Latent Schedule as a Learned Gating Policy}
\label{sec:method}
\begin{figure}[t]
    \centering
    \includegraphics[width=0.98\linewidth]{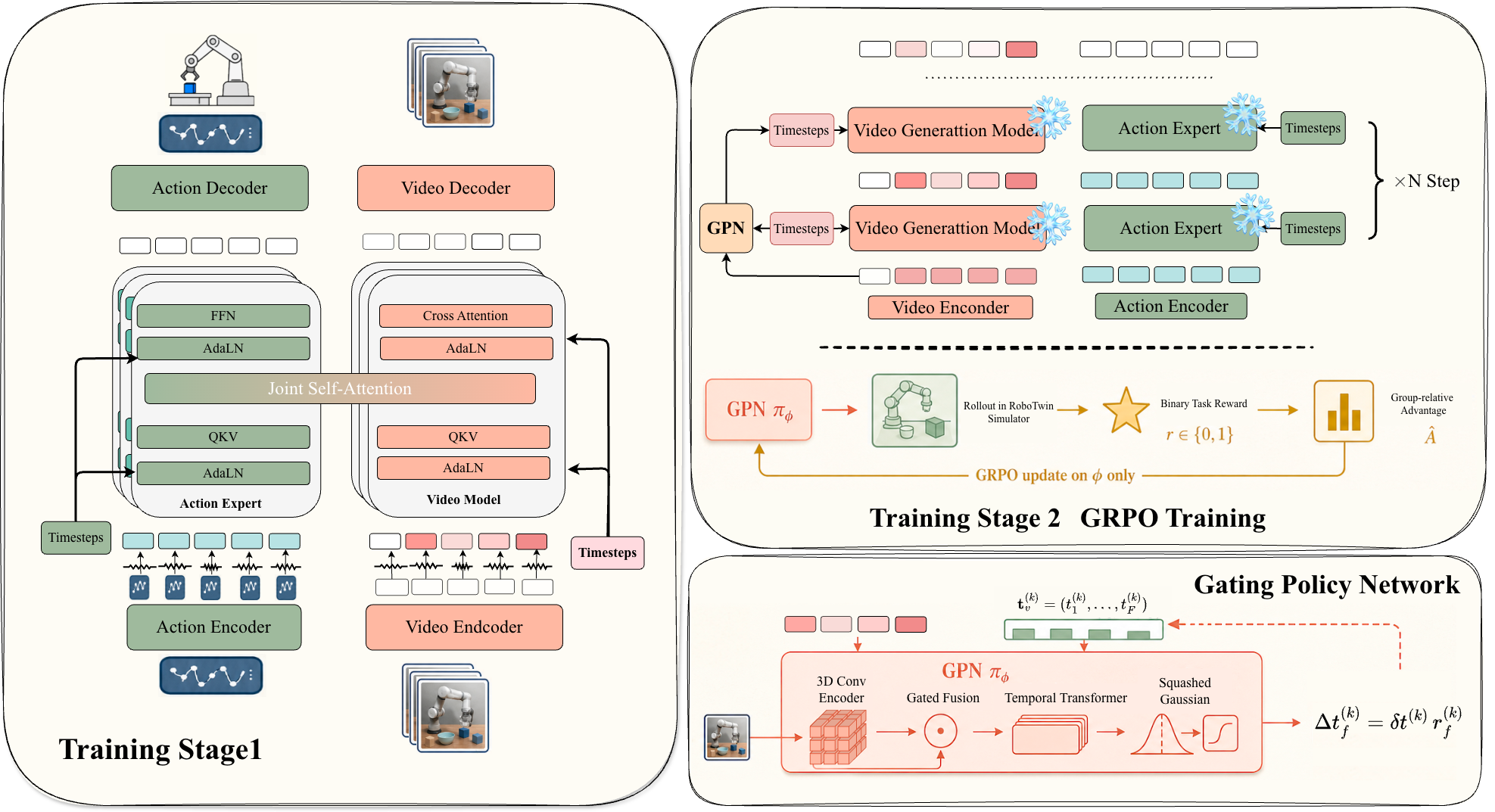}
    \caption{\textbf{Overview of \methodname{}.} The unified framework figure summarizes both the joint-sequence MoT backbone and the Gating Policy Network (GPN). At every denoising step, the GPN reads the current predicted-chunk latents and per-latent times, and emits increments $\Delta t_f$ for $\mathbf{v}$. The observation $v_0$ is pinned at $t_0 = 0$, and the action follows its own global schedule.}
    \label{fig:overview}
\end{figure}

\methodname{} turns the per-latent timestep schedule into a first-class learned object. It has three coupled parts: a joint-sequence MoT backbone (\S\ref{subsec:mot}) on which the gating interpretation lives; independent per-latent timestep sampling (\S\ref{subsec:df}) that makes arbitrary per-latent timestep profiles feasible at inference; and the Gating Policy Network (GPN, \S\ref{subsec:gpn}) trained with GRPO (\S\ref{subsec:grpo}) to emit per-latent time increments $\Delta t_f$ at every denoising step. Operationally, these components are trained in two stages: first the WAM backbone is demonstration-finetuned under independent per-latent timesteps, and then the backbone is frozen while the GPN is optimized from task reward. Figure~\ref{fig:overview} gives an overview.

\subsection{World Action Model and MoT Backbone}
\label{subsec:mot}

Following the notation in \S\ref{subsec:prelim}, we instantiate \methodname{} in the joint video--action modeling paradigm of World Action Models (WAMs), as shown in Fig.~\ref{fig:overview}. Rather than modeling an action-only policy $p_\theta(\aseq\mid v_0,l)$, a WAM co-generates future latents and actions under the current observation and language instruction:
\begin{equation}
\label{eq:wam-paradigm}
    p_\theta(\vseq,\aseq \mid v_0,l),
    \qquad
    g_\theta\!\left(\vseq^{\mathbf{t}_v},\aseq^{t_a},\mathbf{t}_v,t_a,v_0,l\right).
\end{equation}
Here $\mathbf{t}_v=(t_1,\ldots,t_F)$ collects per-frame video timesteps, $t_a$ is the action timestep, and superscripts indicate the current noise level. The backbone head $g_\theta$ denotes the per-step prediction target of the generative backbone, e.g., diffusion noise prediction or flow-matching velocity prediction. This formulation ties prediction and control to the same denoising/flow trajectory: the future latents provide an explicit imagined context, while the action tokens are refined in the same process.

The backbone realizes this WAM with a Mixture-of-Transformers (MoT)~\cite{liang2024mixture} architecture, integrating a chunk-bidirectional video DiT (Wan~2.2-TI2V-5B~\cite{wan2025}) with an Action-Expert DiT~\cite{peebles2023scalable}. The clean observation latent $v_0$ is pinned at $t_0=0$, while predicted latent frames $\{v_f^{t_f}\}_{f=1}^{F}$ and action tokens $\mathbf{a}^{t_a}$ are processed within a joint sequence where all modalities share self-attention blocks but utilize modality-specific feed-forward layers. 

As established in \S\ref{subsec:gating}, this shared attention serves as the physical layer for information gating: the individual noise level $t_f$ modulates the reliability of the $f$-th predicted latent frame's contribution to the action tokens' representations. Notably, while each predicted latent frame $v_f$ is assigned a unique $t_f$, the action tokens $\mathbf{a}$ share a single, global $t_a$ at any given step, ensuring the backbone learns to extract features from a heterogeneously-noised video context to predict a consistent action chunk.

\subsection{Training Substrate: Independent Per-Latent Timestep Sampling}
\label{subsec:df}

To enable the backbone to generalize to arbitrary per-frame schedules at inference, Stage 1 trains the joint video--action backbone with a per-latent timestep sampling regime inspired by Diffusion Forcing~\cite{chen2024diffusion}.

For each training sample, we draw independent timesteps $\mathbf{t}_v = (t_1, \ldots, t_F)$ and a separate action timestep $t_a$, construct the heterogeneously noised input $(\mathbf{v}^{\mathbf{t}_v},\mathbf{a}^{t_a})$, and optimize the backbone with the corresponding diffusion noise target or flow-matching velocity target. This objective ensures the model can effectively "read" through varying levels of uncertainty across the video chunk, making the per-frame schedule a controllable degree of freedom for the policy.

\subsection{Gating Policy Network (GPN)}
\label{subsec:gpn}

\textbf{Policy interface.} The GPN $\pi_\phi$ makes the per-latent schedule a learned control variable. At denoising step $k$, let $\mathbf{t}_v^{(k)}=(t_1^{(k)},\ldots,t_F^{(k)})$ denote the current timesteps of the predicted video latents. The policy observes the heterogeneously noised video chunk $\vseq^{\mathbf{t}_v^{(k)}}$, the clean observation latent $v_0$, and $\mathbf{t}_v^{(k)}$, and outputs timestep decrements only for the predicted video latents:
\begin{equation}
    \Delta\mathbf{t}_v^{(k)}
    =
    \pi_\phi\!\left(
        \vseq^{\mathbf{t}_v^{(k)}},\, \mathbf{t}_v^{(k)}\, v_0
    \right),
    \qquad
    \Delta\mathbf{t}_v^{(k)}\in\mathbb{R}_{+}^{F}.
\end{equation}
The observation remains pinned at $t_0\equiv 0$, while the action tokens follow the original global schedule $t_a^{(k)}$ and are not directly controlled by the GPN. This keeps action denoising synchronized for execution while allowing the video context to be selectively unmasked.

\textbf{Relative-scale action.} Rather than predicting unconstrained absolute decrements, the policy is parameterized through bounded relative scales $\mathbf{r}^{(k)}=(r_1^{(k)},\ldots,r_F^{(k)})\in(0,2)^F$. A lightweight spatiotemporal encoder and squashed-Gaussian actor head parameterize this distribution; the layer-wise architecture and log-density are given in Appendix~\ref{app:gpn}. Let $\delta t^{(k)}$ denote the nominal time decrement that the original scalar denoising schedule would take at step $k$. Each component of the decrement vector is
\begin{equation}
    \Delta t_f^{(k)} = \delta t^{(k)}\, r_f^{(k)},
    \qquad f\in\{1,\ldots,F\}.
\end{equation}

This parameterization preserves the global pace of the sampler while letting the policy decide which predicted latents should be denoised faster or slower at each step. It also avoids asking the policy to learn both the absolute magnitude and the relative priority of each frame from sparse rollout reward.

\textbf{Per-latent denoising trajectory.} The predicted-video timestep vector is then updated component-wise:
\begin{equation}
    t_f^{(k+1)} = \max\{0,\,t_f^{(k)} - \Delta t_f^{(k)}\},
    \qquad f\in\{1,\ldots,F\}.
\end{equation}
We impose no monotonicity or hand-crafted shape prior across frames; the schedule is learned end-to-end from task reward via GRPO (\S\ref{subsec:grpo}). Figure~\ref{fig:overview} shows the GPN in the full framework, and Appendix~\ref{app:gpn} gives the inference algorithm.

\subsection{Training with GRPO}
\label{subsec:grpo}

In Stage 2, we freeze the trained WAM backbone and optimize only the scheduler policy $\pi_\phi$ with GRPO~\cite{shao2024deepseekmath}, which replaces a learned value network by a group-relative baseline. The simulator provides a
binary episodic reward $r\in\{0,1\}$. For a group of $G$ trajectories sampled
under a frozen snapshot $\pi_{\phi_\mathrm{old}}$, the advantage is
\begin{equation}
    \hat{A}_i \;=\; \frac{r_i - \bar{r}}{\mathrm{std}(\{r_j\}_{j=1}^{G}) + \epsilon},
    \qquad \bar{r} \;=\; \tfrac{1}{G}\textstyle\sum_{j} r_j .
\end{equation}
We then perform $E$ epochs of updates on $\pi_\phi$ over the same batch, using
a per-latent importance ratio
\begin{equation}
    \rho_f \;=\; \frac{\pi_\phi(\Delta t_f \mid \vseq^{\mathbf{t}}, \mathbf{t})}
                       {\pi_{\phi_\mathrm{old}}(\Delta t_f \mid \vseq^{\mathbf{t}}, \mathbf{t})} ,
\end{equation}
and optimize the clipped surrogate with an entropy bonus
\begin{equation}
    \mathcal{L}_\mathrm{GRPO}(\phi) \;=\; -\,\mathbb{E}\!\left[
        \sum_f \min\!\Bigl(\rho_f\,\hat{A},\;
        \mathrm{clip}(\rho_f, 1-\varepsilon, 1+\varepsilon)\,\hat{A}\Bigr)
        \;+\; \beta\,\mathcal{H}[\pi_\phi]
    \right] ,
\end{equation}
where $\mathcal{H}[\pi_\phi]$ is the per-step entropy of the Squashed Gaussian
and $\beta$ is decayed exponentially. We omit a KL term since $\pi_\phi$ is trained from scratch with no meaningful reference policy; ratio clipping and the entropy bonus suffice for stability. The backbones are frozen; only $\phi$ is optimized.

% ── 5. Experiments ────────────────────────────────────────────────────────────
\section{Experiments}
\label{sec:experiments}

\subsection{Setup}

\textbf{Benchmark and evaluation protocol.}
We evaluate on \textbf{RoboTwin}~\cite{chen2025robotwin} under its random-scene condition, where object poses, colors, and backgrounds are randomized at every episode. We report two complementary studies: a scaled comparison against strong  baselines, and a controlled ablation that isolates the effect of per-latent scheduling. All methods within each study use the same task list, backbone configuration, rollout budget, and 100-episode-per-task success metric; Appendix~\ref{app:robotwin_details} gives the task-selection details and full 50-task context.

\textbf{Backbone configurations.}
All experiments use the same joint video--action WAM/MoT substrate: a Wan~2.2~\cite{wan2025} video DiT coupled with an Action-Expert DiT through shared self-attention. 
We instantiate it in two training-scale configurations: (i) a Motus-derived configuration with the VLM removed and a batch size of 128, used for controlled ablations and diagnostics, and (ii) a scaled Fast-WAM-style configuration with a batch size of 1024, used for the main performance comparison.
This treats Motus-derived and Fast-WAM-style runs as two configurations of the same WAM/MoT family rather than different architectural claims.

\textbf{Training protocol.}
Following the two-stage procedure introduced in \S\ref{sec:method}, Stage 1 produces a demonstration-trained WAM backbone, denoted \textbf{Stage-1 WAM}, using the official RoboTwin training data. Stage 2 freezes this backbone and trains only the GPN with GRPO. This protocol lets the main comparison, Stage-1 WAM vs.\ \methodname{}, isolate the effect of learning the per-latent denoising schedule on top of the same video--action backbone.

\textbf{Baselines.} The main comparison evaluates \methodname{} against Stage-1 WAM and representative RoboTwin baselines: Fast-WAM, LingBot-VA, $\pi_{0.5}$, and Motus. The ablation compares against Shared-$t$, Stage-1 WAM, and Hand-crafted schedule variants that progressively remove the learned schedule policy.

\textbf{Metric and implementation.} We report task success rate averaged over 100 episodes per task under the random-scene condition of RoboTwin. Full GRPO hyperparameters are given in Appendix~\ref{app:hparams}, and detailed RoboTwin tables are given in Appendix~\ref{app:robotwin_details}.

\subsection{Main Results}

\begin{table*}[t]
  \centering
  \caption{\textbf{Primary RoboTwin random-scene comparison.} Success rates are percentages over 100 episodes per task. Stage-1 WAM is the scaled WAM backbone after demonstration fine-tuning; \methodname{} is the Stage-2 model after GRPO training of the per-latent gating policy. Best per row in bold; $\Delta$ denotes the absolute improvement of \methodname{} over Stage-1 WAM.}
  \label{tab:main}
  \scriptsize
  \setlength{\tabcolsep}{3pt}
  \begin{tabular}{lcccc|ccc}
    \toprule
    \textbf{Task} & \textbf{Fast-WAM} & \textbf{LingBot-VA} & \textbf{$\pi_{0.5}$} & \textbf{Motus} & \textbf{Stage-1 WAM} & \textbf{\methodname{} (ours)} & \textbf{$\Delta$} \\
    \midrule
    Adjust Bottle           & \textbf{100} & 94 & 99 & 93 & \textbf{100} & \textbf{100} & 0 \\
    Beat Block Hammer       & 97 & \textbf{98} & 93 & 88 & \textbf{98} & \textbf{98} & 0 \\
    Blocks Ranking RGB      & \textbf{100} & 98 & 85 & 97 & 97 & 99 & +2 \\
    Blocks Ranking Size     & \textbf{98} & 96 & 26 & 63 & 80 & 87 & +7 \\
    Dump Bin Bigbin         & 96 & 96 & 97 & 91 & 94 & \textbf{99} & +5 \\
    Handover Block          & 81 & 78 & 57 & 73 & 81 & \textbf{84} & +3 \\
    Hanging Mug             & 62 & 28 & 17 & 38 & 60 & \textbf{69} & +9 \\
    Move Can Pot            & 88 & 97 & 55 & 74 & \textbf{100} & \textbf{100} & 0 \\
    Pick Diverse Bottles    & 85 & 82 & 71 & \textbf{91} & 87 & 89 & +2 \\
    Pick Dual Bottles       & 96 & 99 & 63 & 90 & 99 & \textbf{100} & +1 \\
    Place A2B Left          & 93 & 93 & 82 & 79 & 97 & \textbf{100} & +3 \\
    Place A2B Right         & \textbf{99} & 95 & 84 & 87 & 94 & 98 & +4 \\
    Place Bread Basket      & 93 & 95 & 64 & 94 & 97 & \textbf{98} & +1 \\
    Place Bread Skillet     & 93 & 90 & 66 & 83 & 90 & \textbf{96} & +6 \\
    Place Burger Fries      & 99 & 95 & 87 & 98 & 96 & \textbf{100} & +4 \\
    \multicolumn{8}{c}{$\cdots$} \\
    Place Cans Plasticbox   & 96 & 99 & 84 & 94 & 97 & \textbf{100} & +3 \\
    Place Container Plate   & \textbf{100} & 97 & 95 & 99 & 99 & \textbf{100} & +1 \\
    Place Dual Shoes        & 88 & 89 & 75 & 87 & \textbf{94} & \textbf{94} & 0 \\
    Place Fan               & 96 & 93 & 85 & 87 & 95 & \textbf{98} & +3 \\
    Place Mouse Pad         & 89 & \textbf{96} & 39 & 68 & 94 & 94 & 0 \\
    Place Object Basket     & \textbf{88} & \textbf{88} & 76 & 87 & 76 & 79 & +3 \\
    Place Object Scale      & 97 & 95 & 80 & 85 & \textbf{99} & 98 & -1 \\
    Place Object Stand      & 94 & 96 & 85 & 97 & 96 & \textbf{100} & +4 \\
    Press Stapler           & 97 & 82 & 83 & \textbf{98} & 94 & 96 & +2 \\
    Put Bottles Dustbin     & 90 & 91 & 79 & 79 & 90 & \textbf{93} & +3 \\
    Put Object Cabinet      & 89 & 87 & 79 & 71 & 87 & \textbf{94} & +7 \\
    Rotate QRcode           & 89 & \textbf{91} & 87 & 73 & 86 & 86 & 0 \\
    Stack Blocks Three      & 97 & \textbf{98} & 76 & 95 & 96 & \textbf{98} & +2 \\
    Stack Bowls Three       & 81 & 83 & 71 & 87 & \textbf{89} & \textbf{89} & 0 \\
    Turn Switch             & 59 & 45 & 54 & \textbf{78} & 76 & \textbf{78} & +2 \\
    \midrule
    \textbf{Average}        & 91.78 & 91.50 & 76.76 & 87.02 & 92.58 & \textbf{94.28} & 1.70 \\
    \bottomrule
  \end{tabular}
\end{table*}

Table~\ref{tab:main} gives the primary RoboTwin random-scene comparison, with the omitted rows expanded in Appendix~\ref{app:robotwin_details}. The rightmost block separates the demonstration-tuned Stage-1 WAM from the GRPO-trained \methodname{} policy and reports the absolute gain $\Delta$ for each task. Across the displayed tasks, \methodname{} usually preserves already-saturated behavior while improving tasks with remaining headroom: Hanging Mug, Blocks Ranking Size, Put Object Cabinet, and Place Bread Skillet gain 6--9 points, and many medium-difficulty placement and manipulation tasks gain another 2--5 points. The few zero-gain rows are mostly tasks where Stage-1 WAM is already near the ceiling, while the only small regressions shown are on Place Object Scale and Open Microwave. Relative to the prior RoboTwin baselines, this pattern indicates that the learned per-latent schedule is not merely increasing average performance uniformly; it selectively helps tasks where action generation benefits from task-adaptive control over which predicted video latents are made reliable during denoising.

\subsection{Ablation Studies}

\begin{table}[t]
  \centering
  \caption{\textbf{Schedule ablation summary} on the controlled RoboTwin random-scene study (overall average success rate). Each row removes one component of \methodname{}.}
  \label{tab:ablation}
  \small
  \begin{tabular}{lc}
    \toprule
    \textbf{Configuration} & \textbf{Overall Success (\%)} \\
    \midrule
    \methodname{} (full)                                  & \textbf{67.5} \\
    \quad w/o GPN, hand-crafted per-latent timestep schedule       & 63.4 \\
    \quad w/o GPN (Stage-1 WAM)              & 61.3 \\
    \quad w/o GPN, w/o independent per-latent noise training (Shared-$t$) & 57.5 \\
    \bottomrule
  \end{tabular}
\end{table}

Table~\ref{tab:ablation} isolates the role of the schedule design; the per-task breakdown is deferred to Appendix~\ref{app:robotwin_details}, Table~\ref{tab:schedule_ablation_detail}. Removing both the GPN and independent-noise training (Shared-$t$) causes the largest drop (67.5 $\to$ 57.5), confirming that the per-latent timestep schedule---training substrate plus learned gating---is the primary driver. Keeping the Stage-1 WAM setting (independent per-latent noise training with shared-scalar inference) recovers most of the substrate gain (61.3) but leaves the learned schedule unexploited. Replacing the GPN with a hand-crafted monotone timestep schedule adds a further 2.1 points (63.4), showing that \emph{some} per-latent structure is beneficial---but an arbitrary fixed shape is not sufficient; the schedule must be \emph{task-adaptive} to realize the full gain. Only the full \methodname{} closes the remaining gap to 67.5\%, consistent with our framing of the per-latent timestep schedule as a learnable information-gating policy rather than a fixed prior.

\subsection{Qualitative Analysis: Gating, Schedules, and Noise-as-Masking}
\label{subsec:qualitative}

The primary comparison and schedule ablations establish that learning the per-latent timestep schedule helps; we now use two diagnostic views to explain \emph{why} it helps. First, we verify the mechanism predicted by the noise-as-masking interpretation: a noisier latent frame contributes less reliable evidence to the action tokens. Second, we visualize the actual timestep trajectories selected by the GPN for individual chunks from two different tasks, showing that the learned schedule is non-uniform and task-dependent rather than a fixed shared-scalar trajectory. Both probes are taken on the trained \methodname{} policy under the random-scene evaluation protocol.

\begin{figure}[t]
  \centering
  \begin{minipage}[t]{0.33\linewidth}
    \centering
    \includegraphics[width=\linewidth]{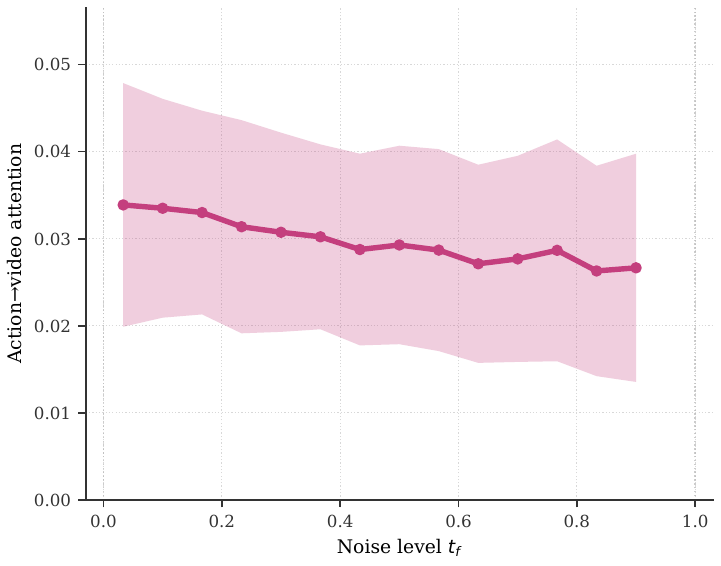}
    \captionof{figure}{\textbf{Noise as masking in the joint self-attention.} Mean action$\!\to\!$video attention versus each predicted frame's current noise level $t_f$. The monotone decay confirms that $t_f$ empirically attenuates each frame's K/V contribution to the action tokens.}
    \label{fig:attn_vs_noise}
  \end{minipage}
  \hfill
  \begin{minipage}[t]{0.64\linewidth}
    \centering
    \includegraphics[width=\linewidth]{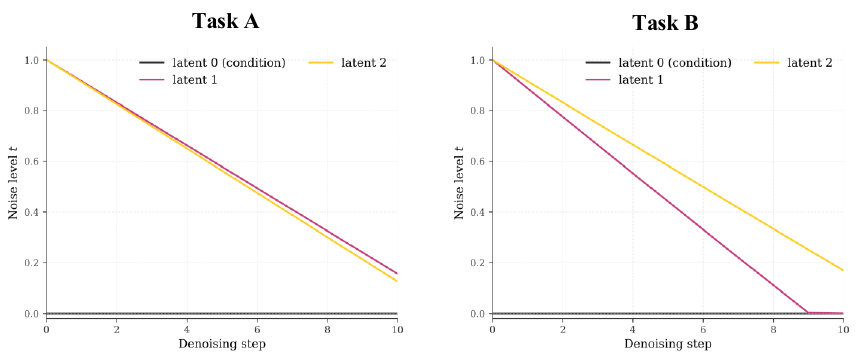}
    \captionof{figure}{\textbf{Task-specific timestep schedules.} Per-frame timestep trajectories for one denoising chunk from two representative tasks. The observation latent is fixed at $t_0{=}0$, while the predicted latents follow different task-dependent trajectories. Under the noise-as-masking interpretation, these curves show how the GPN decides which future latents to expose early to the action tokens and which to keep partially masked.}
    \label{fig:two_task_schedule}
  \end{minipage}
\end{figure}

\paragraph{Noise level acts as a reliability gate.} The noise-as-masking view of Diffusion Forcing~\cite{chen2024diffusion} predicts that, inside the shared self-attention, the noise level $t_f$ on a video token should monotonically attenuate the contribution of its Key/Value projections to the action representations. We test this directly by recording the mean attention from action tokens to each latent frame at every probe step and binning it against that frame's current $t_f$. As shown in Fig.~\ref{fig:attn_vs_noise}, attention scores exhibit a clear monotone decay, so the per-latent vector $\mathbf{t}$ functions empirically---not just nominally---as a bank of continuous reliability gates.

\paragraph{The GPN uses those gates through task-specific schedules.} The attention probe only shows that noise \emph{can} serve as a gate; the scheduler is useful only if it learns when to open or close those gates. Fig.~\ref{fig:two_task_schedule} therefore visualizes the full timestep trajectory of a single denoising chunk from two different tasks. This view exposes the policy decision made at every denoising step: the predicted latents separate into distinct trajectories instead of following one shared curve. Interpreted through Fig.~\ref{fig:attn_vs_noise}, a faster drop in $t_f$ means the corresponding future frame is made visible to the action tokens earlier, whereas a slower drop keeps that frame partially masked and prevents unreliable K/V features from dominating the action update. The important point is not merely that the curves are non-identical, but that their ordering and separation change across tasks. This rules out both the shared-scalar sampler, which has no per-frame degree of freedom, and a fixed monotone hand-crafted schedule, which would impose the same ordering on every chunk. The learned GPN instead allocates the denoising budget according to the current task context, selectively trusting the future latents that are useful for the action while suppressing those that remain ambiguous.

\begin{figure}
    \centering
    \includegraphics[width=0.8\linewidth]{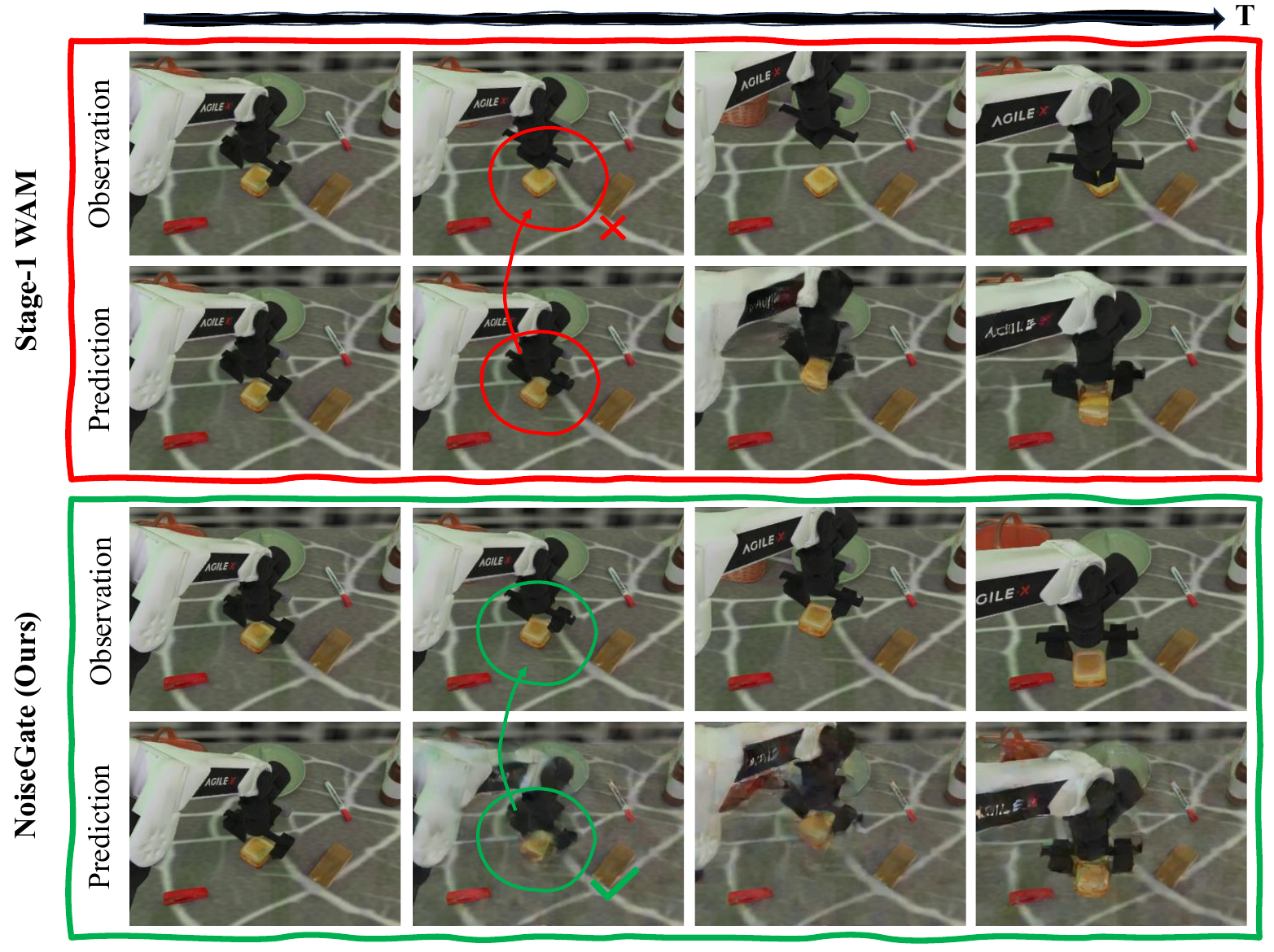}
    \caption{\textbf{Visualization of a real test case comparing Stage-1 WAM and \methodname{}}. For each method, the top row shows the true frames after executing the predicted actions, and the bottom row shows the predicted frames. Stage-1 WAM's standard denoising leads to overconfident grasp prediction (second frame) and premature failure, whereas \methodname{} maintains higher uncertainty in the grasp frame through the learnable scheduler, leveraging other predicted frames to achieve successful grasping.}
    \label{fig:visualization}
\end{figure}

Together, these probes connect the controlled ablations in Table~\ref{tab:ablation} to the mechanism of \methodname{}: independent per-latent noise training makes heterogeneous timestep profiles meaningful, and the GPN learns to choose those profiles so that predicted latent frames are unmasked only to the extent that they are useful for action generation.

\subsection{Case Study}
Finally, Fig.~\ref{fig:visualization} presents a representative test case illustrating how the learned scheduler improves action generation over Stage-1 WAM. Each method is shown with two rows: executed observations on top and predicted future frames on the bottom. Stage-1 WAM produces visually clean future predictions, but appears overconfident about the grasp outcome in the second predicted frame, which leads to a premature grasp and failure. In contrast, \methodname{} keeps the critical grasp frame less fully denoised, reflecting higher uncertainty about the contact event. This uncertainty prevents the action tokens from over-relying on an unreliable future latent; instead, the model integrates evidence from the remaining predicted frames and the current observation, enabling a successful grasp at the critical moment.

% ── 6. Discussion ─────────────────────────────────────────────────────────────
\section{Conclusion}

% \textbf{Conclusion.} 
We presented \methodname{}, which reframes the per-latent timestep schedule of a World Action Model as a learnable information-gating policy and optimizes it against task reward. The approach combines independent per-latent noise sampling during training (from Diffusion Forcing, applied to a chunk-bidirectional backbone), a lightweight GPN emitting per-latent time increments at every step, and GRPO grounded in task success, without hand-crafted shape priors. On RoboTwin random-scene evaluation, \methodname{} improves a strong Stage-1 WAM on the primary comparison and yields a +10.0 point gain over the shared-$t$ baseline in controlled schedule ablations, supporting the view that the per-latent timestep schedule in a WAM is a first-class design object rather than an implicit hyperparameter.

%%%%%%%%%%%%%%%%%%%%%%%%%%%%%%%%%%%%%%%%%%%%%%%%%%%%%%%%%%%%
\bibliography{mybib}
\bibliographystyle{unsrt}

\clearpage

\appendix

\section{Limitations}
\label{app:limitations}

The main limitation of the current schedule-policy training is data-collection efficiency in simulation. GRPO uses sparse task-success rewards, so each update depends on simulator rollouts to obtain learning signal; rollout collection remains the dominant cost and can slow convergence when successful trajectories are rare. Learning a general schedule policy that scales across large-scale multi-task suites is therefore left to future work.

\section{Broader applicability}
\label{app:broader}
The view of per-latent noise as a learnable information-gating policy is not specific to robotics: any generative setting in which a downstream consumer reads, via shared attention, from jointly-denoised tokens with heterogeneous utility---video prediction, autonomous driving, embodied navigation---can in principle be framed the same way.

\section{GPN Architecture and Inference Details}
\label{app:gpn}

This appendix specifies the Gating Policy Network (GPN) used in \S\ref{subsec:gpn}. The goal is to make explicit how the policy maps the current heterogeneously noised video chunk and timestep vector to per-latent scheduling actions.

\paragraph{Interface and notation.}
Let $F'{=}F{+}1$ denote the observation latent together with the $F$ predicted latents. At denoising step $k$, the GPN consumes the current latent stack $\vseq^{\mathbf{t}^{(k)}}\in\mathbb{R}^{B\times C\times F'\times H\times W}$ and per-latent video times $\mathbf{t}^{(k)}\in\mathbb{R}^{B\times F'}$, with $t_0^{(k)}=0$. It returns a sampled relative scale $\mathbf{r}^{(k)}\in(0,1)^F$, the corresponding updated video times $\mathbf{t}^{(k+1)}$, and the sampled-action log-probability used by GRPO. The action timestep $t_a^{(k)}$ is not an input to the GPN policy head; it is advanced by the fixed global schedule.

\paragraph{Network architecture.}
The network first converts the latent stack into one token per video latent. Two strided 3-D convolutions with GroupNorm and SiLU reduce the spatial resolution, producing $\mathbf{z}\in\mathbb{R}^{B\times F'\times C'\times H'\times W'}$. For each latent, attention pooling, average pooling, and max pooling over $(H',W')$ are concatenated and projected to a $D{=}256$ dimensional token, yielding $\mathbf{x}\in\mathbb{R}^{B\times F'\times D}$.

The observation token $\mathbf{x}_0$ is used only as conditioning. Predicted-frame tokens $\mathbf{x}_{1:F}$ are first reweighted by a channel-attention MLP, then fused with $\mathbf{x}_0$ through a gated residual update:
\begin{equation}
    \mathbf{f}_f \;\leftarrow\; \mathbf{f}_f \;+\; \sigma\!\bigl(W_g\,[\mathbf{f}_f;\,\mathbf{x}_0]\bigr) \,\odot\, W_v\,\mathbf{x}_0 .
    \label{eq:gpn-gate}
\end{equation}
Here $W_g\in\mathbb{R}^{D\times 2D}$ produces the sigmoid gate from $[\mathbf{f}_f;\mathbf{x}_0]$, and $W_v\in\mathbb{R}^{D\times D}$ projects the observation token before the residual add.

The resulting predicted-frame tokens receive a learnable frame-position embedding and pass through a three-layer pre-norm Transformer encoder with eight attention heads and GELU feed-forward blocks of width $4D$. In parallel, the current times $\mathbf{t}_{1:F}^{(k)}$ are embedded with sinusoidal features and a two-layer MLP. The latent token and time embedding for each frame are concatenated, normalized, and projected to produce $\mathbf{h}_f\in\mathbb{R}^{D}$. An attention-pooling head forms a global summary $\mathbf{g}=\sum_f w_f\mathbf{h}_f$, where $w_f=\mathrm{softmax}_f(\mathrm{MLP}_\mathrm{score}(\mathbf{h}_f))$.

\paragraph{Squashed-Gaussian policy.}
The actor is a three-layer MLP applied to $[\mathbf{g};\mathbf{h}_1;\ldots;\mathbf{h}_F]$ and outputs $\boldsymbol{\mu}\in\mathbb{R}^F$. The log standard deviation $\log\boldsymbol{\sigma}\in\mathbb{R}^F$ is a learned parameter clamped to $[-5,2]$. For each predicted latent, the policy samples a pre-squash variable
\[
    u_f \sim \mathcal{N}(\mu_f,\sigma_f^2), \qquad r_f=2\sigma(u_f).
\]
Equivalently, for an observed relative-scale action $r_f\in(0,2)$ with $u_f=\mathrm{logit}(r_f)$, the corresponding per-latent log-density is
\begin{equation}
    \log \pi_\phi(r_f \mid s^{(k)}) \;=\; \log \mathcal{N}(u_f;\, \mu_f, \sigma_f^2) \;-\; \bigl(u_f - 2\,\mathrm{softplus}(u_f)\bigr),
    \label{eq:gpn-logprob}
\end{equation}
where $s^{(k)}$ denotes the GPN input state $(\vseq^{\mathbf{t}^{(k)}},\mathbf{t}^{(k)})$. GRPO stores the log-probability of the sampled pre-clipping relative-scale action, $\sum_{f=1}^{F}\log\pi_\phi(r_f\mid s^{(k)})$. Since $\Delta t_f^{(k)}=   \delta t^{(k)}r_f$ differs from $r_f$ only by a fixed step-dependent scale, the additional log-Jacobian term cancels in the new/old policy ratio for a fixed denoising step.

\paragraph{Inference loop.}
Algorithm~\ref{alg:gpn} summarizes how the GPN is invoked inside one denoising step. The algorithm is intentionally written at the scheduling level; the architectural mapping from latents and times to $(\boldsymbol{\mu},\boldsymbol{\sigma})$ is the network described above.

\begin{algorithm}[h]
\caption{Per-step GPN scheduling inside the denoising loop}
\label{alg:gpn}
\begin{algorithmic}[1]
\Require Current latents $\vseq^{\mathbf{t}^{(k)}}$; video times $\mathbf{t}^{(k)}$ with $t_0^{(k)}=0$; scalar-schedule decrement $\delta t^{(k)}$; action time $t_a^{(k)}$.
\Ensure Updated video times $\mathbf{t}^{(k+1)}$; next action time $t_a^{(k+1)}$; sampled-action log-probability.
\State Encode $(\vseq^{\mathbf{t}^{(k)}},\mathbf{t}^{(k)})$ with the GPN to obtain $(\boldsymbol{\mu},\boldsymbol{\sigma})$.
\State Sample $u_f\sim\mathcal{N}(\mu_f,\sigma_f^2)$ and set $r_f\gets\sigma(u_f)$ for $f=1,\ldots,F$.
\State Set $\Delta t_f^{(k)}\gets 2\,\delta t^{(k)}r_f$ for $f=1,\ldots,F$.
\State Update $t_f^{(k+1)}\gets\max(0,t_f^{(k)}-\Delta t_f^{(k)})$ for $f=1,\ldots,F$; keep $t_0^{(k+1)}\gets 0$.
\State Advance $t_a^{(k+1)}$ using the fixed global action schedule.
\State Run the frozen MoT backbone with $(\mathbf{t}^{(k+1)},t_a^{(k+1)})$ to produce the next denoising state.
\State Store $\sum_{f=1}^{F}\log\pi_\phi(r_f\mid s^{(k)})$ for the GRPO update.
\end{algorithmic}
\end{algorithm}

During GRPO training, only the GPN parameters $\phi$ are optimized. The Wan~2.2 video DiT and Action-Expert DiT backbone remain frozen throughout the rollout collection and policy update.

\section{GRPO Hyperparameters}
\label{app:hparams}

Table~\ref{tab:hparams} lists the GRPO training hyperparameters used in all experiments.

\begin{table}[h]
  \centering
  \caption{\textbf{GRPO hyperparameters.}}
  \label{tab:hparams}
  \small
  \begin{tabular}{lc}
    \toprule
    \textbf{Hyperparameter} & \textbf{Value} \\
    \midrule
    Group size $G$ & 8 \\
    Clip range $\varepsilon$ & 0.2 \\
    Inner epochs $E$ & 4 \\
    Optimizer & AdamW \\
    Learning rate & $1\times10^{-4}$ \\
    Gradient-norm clip & 1.0 \\
    Entropy coefficient $\beta_0$ & 0.01 \\
    Entropy decay rate & 0.999 per step (floored at 0) \\
    Denoising steps $K$ & 10 \\
    Training epochs & 50 \\
    Episodes per epoch & 8 \\
    Rollout workers (GPUs) & 8 \\
    Seed & 42 \\
    \bottomrule
  \end{tabular}
\end{table}

\section{Detailed RoboTwin Results}
\label{app:robotwin_details}

The controlled ablation uses the Motus-derived batch-128 configuration to support exhaustive component comparisons and diagnostics; the primary comparison uses the scaled Fast-WAM-style batch-1024 configuration. Table~\ref{tab:schedule_ablation_detail} gives the per-task breakdown behind the schedule ablation summary in Table~\ref{tab:ablation}. Table~\ref{tab:robotwin_full} provides the full 50-task RoboTwin random-scene context. Unavailable full-method entries are marked with ``--''.

\begin{table}[h]
  \centering
  \caption{\textbf{Per-task schedule ablation on RoboTwin random-scene tasks (100 episodes each).} All variants in this table use the same controlled Motus-derived backbone configuration and evaluation protocol. ``Shared-$t$'' is the shared-scalar baseline; ``Stage-1 WAM'' uses independent per-latent noise training but denoises with a shared scalar at inference; ``Hand-crafted'' uses a fixed monotone per-latent timestep schedule; ``\methodname{}'' learns the per-latent timestep schedule via GRPO. Best per row in bold.}
  \label{tab:schedule_ablation_detail}
  \small
  \setlength{\tabcolsep}{5pt}
  \begin{tabular}{lcccc}
    \toprule
    \textbf{Task} & \textbf{Shared-$t$} & \textbf{Stage-1 WAM} & \textbf{Hand-crafted} & \textbf{\methodname{} (Ours)} \\
    \midrule
    Adjust Bottle             & 0.88 & 0.85 & 0.85 & \textbf{0.94} \\
    Open Microwave            & 0.87 & 0.86 & 0.84 & \textbf{0.91} \\
    Move Playingcard Away     & 0.86 & 0.87 & 0.90 & \textbf{0.91} \\
    Place Bread Basket        & 0.75 & 0.82 & 0.87 & \textbf{0.92} \\
    Place A2B Right           & 0.75 & 0.68 & 0.70 & \textbf{0.78} \\
    Beat Block Hammer         & 0.76 & 0.70 & 0.73 & \textbf{0.82} \\
    Blocks Ranking RGB        & 0.72 & 0.76 & 0.73 & \textbf{0.82} \\
    Place Fan                 & 0.56 & 0.57 & 0.60 & \textbf{0.62} \\
    Lift Pot                  & 0.51 & 0.56 & 0.51 & \textbf{0.57} \\
    Place Mouse Pad           & 0.50 & 0.49 & 0.52 & \textbf{0.54} \\
    Place Dual Shoes          & 0.36 & 0.46 & 0.51 & \textbf{0.59} \\
    Place Can Basket          & 0.40 & 0.44 & 0.44 & \textbf{0.49} \\
    Move Pillbottle Pad       & 0.32 & 0.47 & 0.50 & \textbf{0.51} \\
    Move Can Pot              & 0.21 & 0.27 & 0.30 & \textbf{0.32} \\
    Pick Diverse Bottles      & 0.11 & 0.34 & 0.39 & \textbf{0.39} \\
    \midrule
    \textbf{Average}          & 0.575 & 0.613 & 0.634 & \textbf{0.675} \\
    \bottomrule
  \end{tabular}
\end{table}

\begin{table*}[t]
  \centering
  \tiny
  \caption{\textbf{Full 50-task RoboTwin random-scene context.} All entries are success rates in percent. The Fast-WAM-family and prior-method columns are adapted from the corresponding RoboTwin detail table; the Stage-1 WAM column uses our 50-task random-scene evaluation. ``--'' denotes full-method entries not reported.}
  \label{tab:robotwin_full}
  \setlength{\tabcolsep}{2pt}
  \resizebox{\textwidth}{!}{
  \begin{tabular}{lcccccccc}
    \toprule
    \textbf{Task} & \textbf{Fast-WAM} & \textbf{Fast-WAM-Joint} & \textbf{Fast-WAM-IDM} & \textbf{LingBot-VA} & \textbf{$\pi_{0.5}$} & \textbf{Motus} & \textbf{Stage-1 WAM} & \textbf{\methodname{}} \\
    \midrule
    Adjust Bottle & 100 & 99 & 99 & 94 & 99 & 93 & 100 & 100 \\
    Beat Block Hammer & 97 & 98 & 98 & 98 & 93 & 88 & 98 & 98 \\
    Blocks Ranking RGB & 100 & 100 & 99 & 98 & 85 & 97 & 97 & 99 \\
    Blocks Ranking Size & 98 & 91 & 90 & 96 & 26 & 63 & 80 & 87 \\
    Click Alarmclock & 100 & 100 & 100 & 100 & 89 & 100 & 100 & 100 \\
    Click Bell & 100 & 98 & 96 & 100 & 66 & 100 & 100 & 100 \\
    Dump Bin Bigbin & 96 & 95 & 98 & 96 & 97 & 91 & 94 & 99 \\
    Grab Roller & 100 & 100 & 100 & 100 & 100 & 100 & 100 & 100 \\
    Handover Block & 81 & 91 & 94 & 78 & 57 & 73 & 81 & 84 \\
    Handover Mic & 100 & 100 & 99 & 96 & 97 & 63 & 100 & 100 \\
    Hanging Mug & 62 & 56 & 62 & 28 & 17 & 38 & 60 & 69 \\
    Lift Pot & 100 & 100 & 100 & 99 & 85 & 99 & 99 & 99 \\
    Move Can Pot & 88 & 99 & 100 & 97 & 55 & 74 & 100 & 100 \\
    Move Pillbottle Pad & 99 & 100 & 100 & 99 & 61 & 96 & 100 & 100 \\
    Move Playingcard Away & 100 & 100 & 100 & 99 & 84 & 96 & 100 & 100 \\
    Move Stapler Pad & 64 & 81 & 85 & 79 & 42 & 85 & 81 & 85 \\
    Open Laptop & 100 & 92 & 92 & 94 & 96 & 91 & 99 & 99 \\
    Open Microwave & 45 & 14 & 53 & 86 & 77 & 91 & 77 & 69 \\
    Pick Diverse Bottles & 85 & 87 & 89 & 82 & 71 & 91 & 87 & 89 \\
    Pick Dual Bottles & 96 & 99 & 98 & 99 & 63 & 90 & 99 & 100 \\
    Place A2B Left & 93 & 96 & 96 & 93 & 82 & 79 & 97 & 100 \\
    Place A2B Right & 99 & 95 & 98 & 95 & 84 & 87 & 94 & 98 \\
    Place Bread Basket & 93 & 94 & 97 & 95 & 64 & 94 & 97 & 98 \\
    Place Bread Skillet & 93 & 93 & 95 & 90 & 66 & 83 & 90 & 96 \\
    Place Burger Fries & 99 & 100 & 99 & 95 & 87 & 98 & 96 & 100 \\
    Place Can Basket & 69 & 23 & 28 & 84 & 62 & 76 & 58 & 62 \\
    Place Cans Plasticbox & 96 & 98 & 96 & 99 & 84 & 94 & 97 & 100 \\
    Place Container Plate & 100 & 98 & 96 & 97 & 95 & 99 & 99 & 100 \\
    Place Dual Shoes & 88 & 89 & 87 & 89 & 75 & 87 & 94 & 94 \\
    Place Empty Cup & 100 & 100 & 100 & 100 & 99 & 98 & 100 & 100 \\
    Place Fan & 96 & 96 & 95 & 93 & 85 & 87 & 95 & 98 \\
    Place Mouse Pad & 89 & 91 & 93 & 96 & 39 & 68 & 94 & 94 \\
    Place Object Basket & 88 & 81 & 82 & 88 & 76 & 87 & 76 & 79 \\
    Place Object Scale & 97 & 99 & 99 & 95 & 80 & 85 & 99 & 98 \\
    Place Object Stand & 94 & 98 & 100 & 96 & 85 & 97 & 96 & 100 \\
    Place Phone Stand & 99 & 100 & 99 & 97 & 81 & 86 & 98 & 98 \\
    Place Shoe & 99 & 97 & 98 & 98 & 93 & 97 & 97 & 99 \\
    Press Stapler & 97 & 50 & 57 & 82 & 83 & 98 & 94 & 96 \\
    Put Bottles Dustbin & 90 & 95 & 92 & 91 & 79 & 79 & 90 & 93 \\
    Put Object Cabinet & 89 & 90 & 90 & 87 & 79 & 71 & 87 & 94 \\
    Rotate QRcode & 89 & 92 & 86 & 91 & 87 & 73 & 86 & 86 \\
    Scan Object & 92 & 92 & 90 & 91 & 65 & 66 & 87 & 92 \\
    Shake Bottle & 100 & 100 & 100 & 97 & 97 & 97 & 100 & 100 \\
    Shake Bottle Horizontally & 100 & 100 & 100 & 99 & 99 & 98 & 100 & 100 \\
    Stack Blocks Three & 97 & 97 & 95 & 98 & 76 & 95 & 96 & 98 \\
    Stack Blocks Two & 100 & 100 & 100 & 98 & 100 & 98 & 100 & 100 \\
    Stack Bowls Three & 81 & 86 & 83 & 83 & 71 & 87 & 89 & 89 \\
    Stack Bowls Two & 98 & 95 & 96 & 98 & 96 & 98 & 99 & 100 \\
    Stamp Seal & 94 & 99 & 94 & 97 & 55 & 92 & 96 & 97 \\
    Turn Switch & 59 & 72 & 74 & 45 & 54 & 78 & 76 & 78 \\
    \midrule
    \textbf{Average} & 91.78 & 90.32 & 91.34 & 91.50 & 76.76 & 87.02 & 92.58 & 94.28 \\
    \bottomrule
  \end{tabular}
  }
\end{table*}

\section{Additional Qualitative Analyses}
\label{app:additional}

This appendix collects additional diagnostic views that complement the two qualitative probes in \S\ref{subsec:qualitative}. They provide layer-stratified attention statistics and broader schedule summaries beyond the compact main-text visualization.

\subsection{Mechanism: layer-stratified action$\to$video attention vs.\ noise level}
\label{app:by_layer}

\begin{figure}[h]
  \centering
  \includegraphics[width=0.6\linewidth]{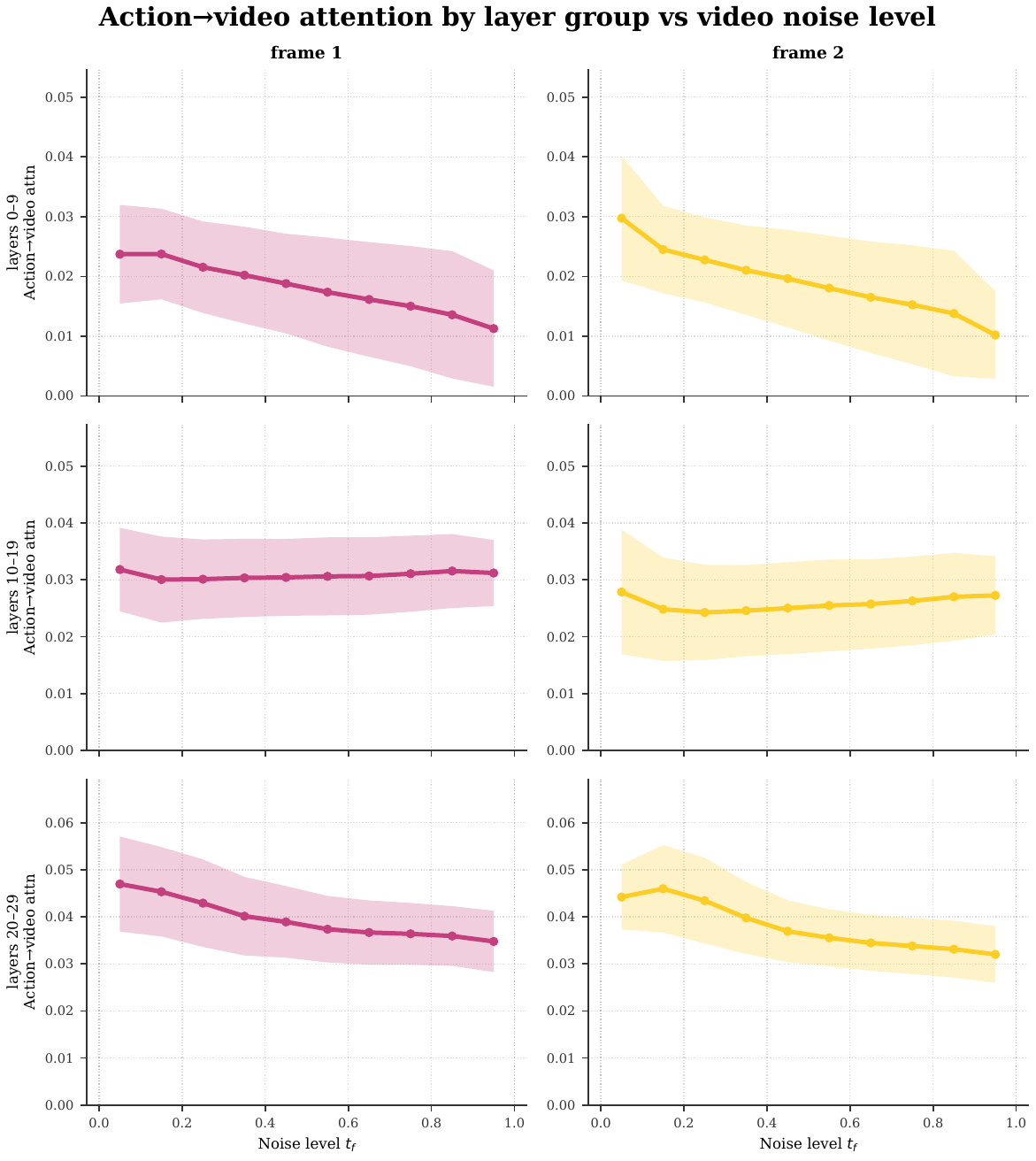}
  \caption{\textbf{Layer-stratified version of Fig.~\ref{fig:attn_vs_noise}.} Same binned-mean $\pm 1$ std curve of action$\to$video attention vs.\ frame noise level $t_f$, separated into three layer groups (early: layers 0--9; middle: 10--19; late: 20--29). The gating effect is concentrated in the early and late layers, with the middle layers behaving near-uniformly in $t_f$.}
  \label{fig:attn_by_layer}
\end{figure}

Fig.~\ref{fig:attn_vs_noise} reports the mechanism aggregated across the recorded probe layers. Fig.~\ref{fig:attn_by_layer} stratifies the same scatter into three layer groups (early / middle / late). The monotone decay of action$\to$video attention with respect to $t_f$ is preserved in the early- and late-layer groups, while the middle-layer group is near-flat. The mechanism is therefore not an artifact of any single probe layer, but it is also non-uniform across depth: the gating effect is strongest where the action tokens directly read out video features (early and late layers), and weakest in the intermediate layers, where representations are already largely modality-agnostic.

\subsection{Cross-task structure of final-step residual noise}
\label{app:final_t_by_task}

\begin{figure}[h]
  \centering
  \includegraphics[width=0.6\linewidth]{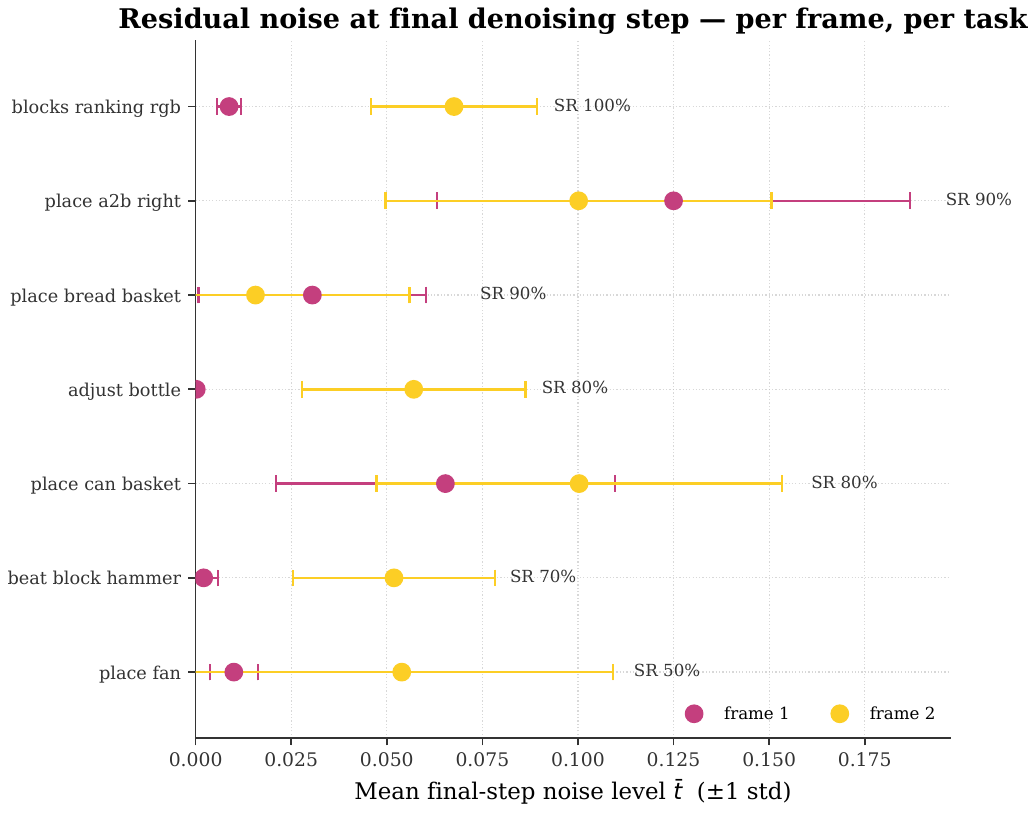}
  \caption{\textbf{Final-step residual noise by task and future-frame index.} For each RoboTwin task, dots show the mean final-step noise level $\bar{t}$ of each predicted future frame over all evaluated chunks, and horizontal bars show $\pm 1$ standard deviation. Tasks are sorted by success rate. The non-uniform, task-dependent residuals show that the learned GPN does not collapse to a shared denoising endpoint; instead, it selectively leaves some future latents partially masked when their contribution should be attenuated.}
  \label{fig:final_t_by_task}
\end{figure}

Fig.~\ref{fig:final_t_by_task} further examines the timestep schedule learned by the GPN from a task-level perspective. For each task, we collect all evaluated chunks and report the mean final-step noise level $\bar{t}$ of each predicted future frame, with horizontal bars denoting $\pm 1$ standard deviation. The conditioning frame is omitted because it is fixed at $t_0 \equiv 0$. Tasks are sorted by their success rate, so the figure jointly shows task difficulty and the residual masking pattern selected by the learned schedule.

The learned policy does not simply drive all predicted frames to the same clean endpoint. Instead, the two future frames exhibit distinct residual-noise levels, and the gap is task-dependent. In several tasks, frame~1 is almost fully denoised while frame~2 retains a larger residual $t$, indicating that the policy chooses to keep the farther future latent partially masked at the end of action generation. The effect is especially visible on tasks such as \texttt{place\_a2b\_right} and \texttt{place\_can\_basket}, where the uncertainty over future contact or placement outcomes is higher. This supports the information-gating interpretation: the GPN learns when a predicted latent should contribute as reliable evidence and when it should remain attenuated rather than forcing every latent through a shared scalar schedule.

\end{document}